\documentclass[10pt,twocolumn,letterpaper]{article}

\usepackage{iccv}
\usepackage{times}
\usepackage{epsfig}
\usepackage{graphicx}
\usepackage{amsmath}
\usepackage{amssymb}
\usepackage{kotex}
\usepackage{capt-of,etoolbox}
\usepackage{subfigure}
\usepackage{multirow}

\usepackage[breaklinks=true,bookmarks=false]{hyperref}

\newcommand\smn[1]{\textcolor{black}{#1}}
\newcommand\knj[1]{\textcolor{black}{#1}}

\newcommand\sm[1]{\textcolor{black}{#1}}
\newcommand\sw[1]{\textcolor{black}{#1}}
\newcommand\hy[1]{\textcolor{black}{#1}}
\newcommand\nj[1]{\textcolor{black}{#1}}

\iccvfinalcopy % *** Uncomment this line for the final submission

\def\iccvPaperID{3919} % *** Enter the ICCV Paper ID here

\ificcvfinal\fi

\begin{document}
%%%%%%%%% TITLE
\title{Sym-parameterized Dynamic Inference for Mixed-Domain Image Translation}

\author{Simyung Chang$^{1,2}$, SeongUk Park$^{1}$, John Yang$^{1}$, Nojun Kwak$^{1}$\\
$^{1}$Seoul National University, Seoul, Korea\\
$^{2}$Samsung Electronics, Suwon, Korea\\
{\tt\small \{timelighter, swpark0703, yjohn, nojunk\}@snu.ac.kr}
}

\twocolumn[{%
\renewcommand\twocolumn[1][]{#1}%
\maketitle
\begin{center}
\vskip -0.25in
    \centering
    \includegraphics[width=1.\textwidth, height=8.2cm]{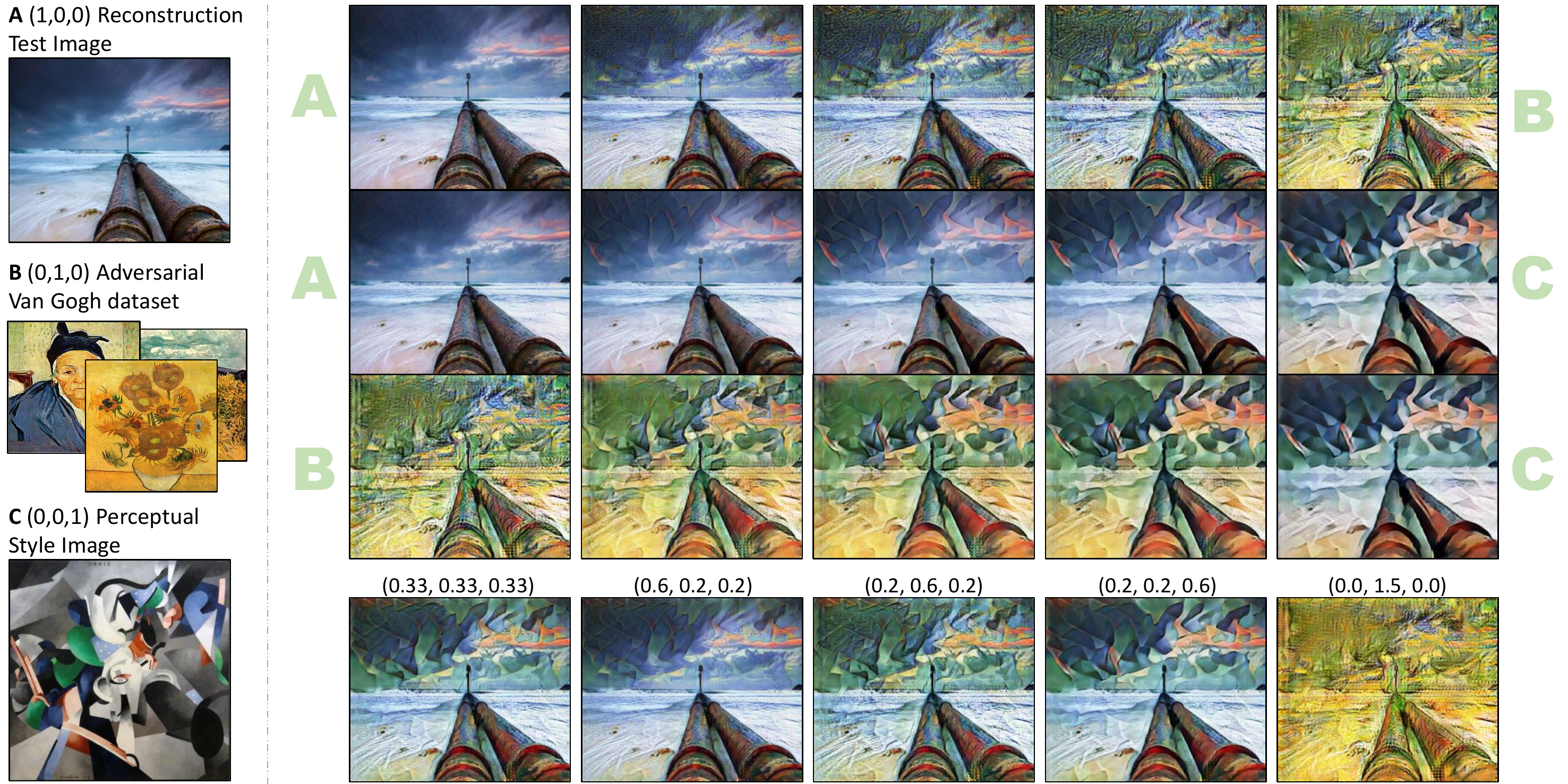}
   \captionof{figure}{Results of image translation to mixed-domains. These images are obtained by learning the SGN for three losses (reconstruction, adversarial, perceptual) and then by inferring the input of a test image in a single generator with only changing the sym-parameter. The numbers in the parentheses are sym-parameters for each A, B, and C domain.}
    \label{fig:main}
\end{center}%
}]

\ificcvfinal\thispagestyle{empty}\fi

%%%%%%%%% ABSTRACT
\begin{abstract}
Recent advances in image-to-image translation have led to some ways to generate multiple domain images through a single network. 
However, there is still a limit in creating an image of a target domain without a dataset on it. 
We propose a method \hy{that expands} the concept of `multi-domain' from data to the loss area \hy{and learns the combined} characteristics of each domain to \hy{dynamically infer translations of}
%create an
images \hy{in mixed domains}. First, we introduce Sym-parameter and its learning method for variously mixed
losses \hy{while synchronizing} them with input conditions. Then, we propose Sym-parameterized Generative Network (SGN) which is
empirically confirmed of learning mixed
characteristics of various data and losses, and
translating images to any mixed-domain without ground truths, such as 30\% Van Gogh and 20\% Monet and 40\% snowy. 
\end{abstract}

%%%%%%%%% BODY TEXT
\section{Introduction}
Recently, literature on multi-domain deep image translation has introduced many methods that learn the joint distribution of two or more domains and find transformations among them.
{Particularly, a} single generator is able to translate images to multiple domains based on training data distributions \cite{choi2017stargan,perarnau2016invertible,yu2018singlegan}.
However, translating style features across domains seen by the model is different from `creativeness'. 
Consider a case of generating a translated image with a style that is 20\% of Van Gogh, 50\% of Picasso and 30\% of the original image. 
Since the ground truths for learning such {a} translation do not exist, 
the target distribution to approximate can not explicitly be provided for conventional deep generative networks. 

If we construe that the optimum of the target style is a weighted sum of optima of the candidate styles, then the objective function can be defined by a weighted sum of objective functions of those.
In this end, if the weights are set as hyper-parameters, they can be preselected and learned to generate images outside of a domain even without ground truths \cite{elgammal2017can}. 
Even in such a case, not only a criterion on selections of the parameters is vague, but every training for cross-domain translations must also be impractically done with a unique set of weights. 
{Therefore, it} would be more efficient to dynamically control them during inferences for desired translations.
We, in this paper, present a concept of \textit{sym-parameters} that enable human users to control them so that influences of candidate domains on final translations can be heuristically adjusted during inferences.

\nj{In our method, a}long with inputs, sym-parameters are inputted as a condition into our proposed generator network, Sym-parameterized Generative Network (SGN). And also, these sym-parameters are synchronously set as weights for the linear combination of multiple loss functions.
With the proposed setting, we have verified that a single network is able to generate corresponding images of a mixed-domain based on an arbitrarily weighted combination of loss functions without direct ground truths.
While an SGN utilizes multiple loss functions that conventional image-to-image translation models use (e.g. reconstruction loss, GAN (adversarial) loss, perceptual loss), the sym-parameter conditions the weights of {these} losses for variously purposed translations. If an SGN, as an exemplar case depicted in Figure \textcolor{red}{1}, uses GAN loss for training {Van Gogh} style and perceptual loss for {\textit{Udnie} of \textit{Francis Picabia}}, sym-parameters allow adjusting the ratio of the styles to create correspondingly styled images. 
Through experiments, {we found that} sym-parameters conditioned within a model in the ways performed in typical conditional methods~\cite{choi2017stargan,lample2017fader,zhu2017toward}, {fail} to yield our intended generations.
To overcome this, we additionally propose Conditional Channel Attention Module (CCAM).

To summarize, our contributions are as follows:\\
(1) We propose the concept of sym-parameter and its learning method that can control the weight between losses during inferences.\\
(2) We introduce SGN, a novel generative network that expands the concept of `multi-domain' from data to the loss area using sym-parameters. \\
(3) Experimental results show that SGN could translate images to mixed-domain without ground truths.

%-------------------------------------------------------------------------
\section{Related work}

\noindent\textbf{Generative Adversarial Networks} 
Recently, Generative Adversarial Networks (GANs) \cite{goodfellow2014generative} have been actively adapted to many image generation tasks \cite{berthelot2017began,karras2017progressive,ledig2017photo,radford2015unsupervised}.
GANs are typically composed of two networks: a generator and a discriminator. 
The discriminator is trained to distinguish the generated samples (fake samples) from the ground-truth images (real samples), 
while the generator learns to generate samples so that the discriminator misjudges.
This training method is called adversarial training which our method uses for both the generator and the discriminator to learn the distribution of {a} real dataset.

\noindent\textbf{Conditional Image Synthesis} 
By conditioning concurrently with inputs, image generation methods learn conditional distribution of a domain.
CVAE \cite{sohn2015learning} uses conditions to assign intentions to VAEs \cite{kingma2013auto}. 
Conditional image generation methods that are based on GANs also have been developed \cite{chen2016infogan,dosovitskiy2017learning,mirza2014conditional, odena2016conditional,perarnau2016invertible,zhang2017stackgan}, using class labels or other characteristics. 
Conditional GANs are also used in domain transfers \cite{kim2017learning,taigman2016unsupervised} and super-resolution \cite{ledig2017photo}. 
While the methods from \cite{choi2017stargan,perarnau2016invertible} use discrete conditioning (\textit{either 0 or 1}), our method uses continuous values for input conditioning.

\noindent\textbf{Image to Image Translation} While there exist generative models that generate images based on sampling (\textit{i.e. GANs, VAEs}), models that generate images {for} given base input images are also studied. They mostly use {a}utoencoders \cite{kramer1991nonlinear} and among them, one of the most representative and recent works is \cite{isola2017image}, which uses adversarial training with conditions. CycleGAN \cite{zhu2017unpaired} and DiscoGAN \cite{kim2017learning} translate either style or domain of input images. Johnson \textit{et al.} \cite{johnson2016perceptual} have proposed perceptual loss in order to train feed-forward network for image style transformation. 
Since most of them use convolutional autoencoders with ResBlocks \cite{he2016deep} or U-Net \cite{ronneberger2015u} structure,
we also utilize the structure of CycleGAN {\cite{zhu2017unpaired}}, but additionally apply sym-parameters to image-to-image translation tasks in {the form of} continuous valued conditions. 
We also adopt the perceptual loss harmonized with the GAN loss and the reconstruction loss terms.

\noindent\textbf{One generator to multiple domains} Many have extended the study of image-to-image translations to multiple domains with a single generator network. 
IcGAN \cite{perarnau2016invertible}, StarGAN \cite{choi2017stargan} and SingleGAN \cite{yu2018singlegan} address the problem of previously reported generative models that they are stuck with two domains, and achieve meaningful results on their extended works by using hard labels of each domain.
Methods regarding image generation problems also have been proposed. ACGAN \cite{odena2016conditional} uses auxiliary classifier to generate images by providing class information as a condition in the input. 
In different perspective, CAN \cite{elgammal2017can} tries generating artworks by blending multiple domains. The method trains the generator of GAN to confuse the auxiliary classifier to judge fake samples in forms of uniform-distribution.
In this study, we make a good use of conditions synchronized with loss functions not only to transfer to multiple domains, but also to mix each styles simultaneously by using expandable loss terms.

\section{Proposed Method}
Our goal is to
learn distributions from multiple domains by varying weighted loss functions in order to dynamically translate images into a mixed-domain. 
In order to control mixing ratio during inferences, 
the corresponding conditions must be inputted and trained with the model. 
For such purpose, we present \textit{sym-parameters} which are symmetrically set inside (as condition inputs) {as well as} outside (as weights of multiple loss functions) of a generator.
The sym-paraemters allow transitive learning without explicit ground truth images among diverse mixtures of multiple domains and loss functions.
The generator of sym-parameterized generative networks (SGN) can thus be controlled during inferences 
unlike conventional generators that infer strictly as optimized for a specific dataset or a particular loss function.

\subsection{Sym-parameter}
\begin{figure}
    \centering
	\includegraphics[width=1.\linewidth]{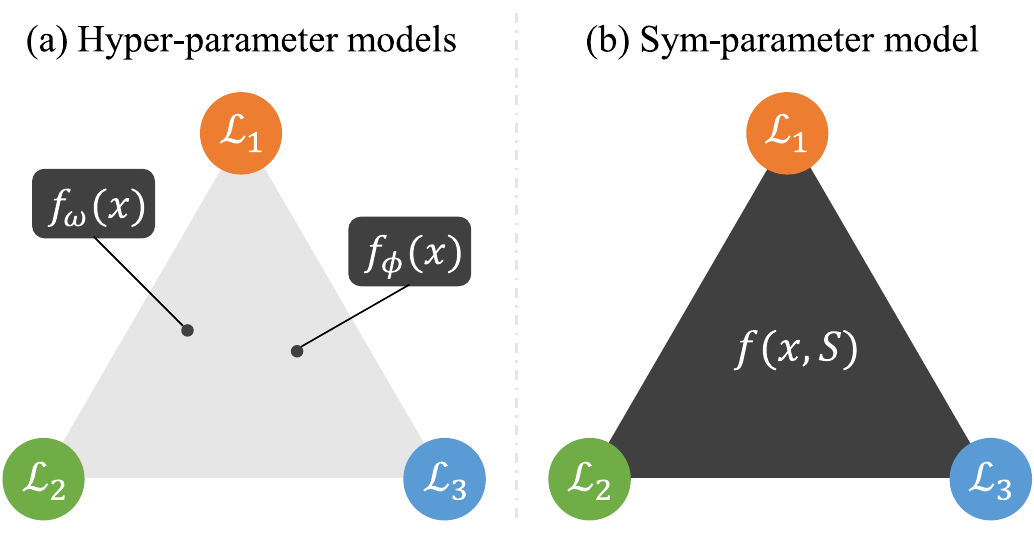}
    \caption{\textbf{The concept of sym-parameter} (a) This model should be learned by changing the hyper-parameters when it is required to use different weights {among} multiple losses. (b) Our proposed method has the effect of modifying the weight {among} losses by changing the sym-parameter $S$ for inference in a single model. $\mathcal{L}_1, \mathcal{L}_2, \mathcal{L}_3$ represent different types of losses. $x$ and $f$ are the input and the output function. \nj{$\omega$ and $\phi$ are the hyper-parameters}}
    \label{fig:sym-param}
\end{figure}

{By trying to find} not only the optimum of each candidate objective function but also {the optima for various combinations of them}, we desire to control the {mixing} weights during inferences. 
We propose human controllable parameters, \textbf{\textit{Sym-parameters}}, that can replace typical hyper-parameters for weighing multiple loss functions. 
As the prefix ``sym-'' is defined in dictionaries as ``with; along with; together; at the same time'', 
sym-parameters are fed into a \sm{model}, \textit{symmetrically} set as weights of the candidate loss functions and \textit{synchronized} after training. 
If $k$ number of different loss functions, {$\mathcal{L}_1, \cdots, \mathcal{L}_k$}, are engaged, then the sym-parameter $S$ is defined as a $k$-dimensional vector $(s_1, \cdots, s_k)$. 
The total loss of a model $f(x,S)$ that takes inputs $x$ and sym-parameters $S$ is:
\begin{align}
\begin{split}\label{eq:sym-param}
&\mathcal{L}(f, S) = s_1 \mathcal{L}_1(f(x,S)) + ... + s_k \mathcal{L}_k(f(x,S)) \\
&\text{where} \sum_{i=1}^{k}{s_i} = 1 \text{ and }  s_i \geq 0 \text{ for all } i \in [1,k]. 
\end{split}
\end{align}
Having the total sum of sym-parameters to be 1, the total loss $\mathcal{L}$ defined by {the} function $f$ and sym-parameters $S$ is a weighted sum of sub-loss functions, each of which {is} weighed by {the corresponding element of $S$}.
\sm{The conventional hyper-parameter model predicts output $y$ using the $\hat{y}=f_{h_t}(x)$ model with hyper-parameter $h_t$. In this case, it is difficult to predict $y$ for $h_{t'}$, that is not used \nj{during training,}
because \nj{the} network $f$ is not a conditional function for \nj{the} hyper-parameter $h_{t'}$ that \nj{was} not used at training time. However, our model uses the weight of losses as input $S$ in the form $\hat{y}=f(x,S)$, and $f$ has conditional output for $S$ as well as $x$. Thus, it can predict $\hat{y}$ for \nj{various combinations}  of losses that \nj{were} used in training.}
Figure~\ref{fig:sym-param} depicts the concept {showing the} difference between a sym-parametrized model and conventional models using multiple loss functions weighted by hyper-parameters.
While a new model is required for learning each combination of weights if conventional methods are used, our method allows a single model to manage various combination of weights {through} one learning.
We, in the experiment section, verify that not only each sub-objective function is optimized but various linear sums of them also are.
For example, if a neural model wants to perform {both the regression and the classification tasks}, 
the optimization is processed towards minimizing the regression loss when $S=(1,0)$, the classification loss when $S=(0,1)$ and the weighted sum of the losses when {$S=(i,j)_{i,j\in\mathbb{R}, i+j=1}$.}

\begin{figure*}
    \centering
	\includegraphics[width=.95\linewidth]{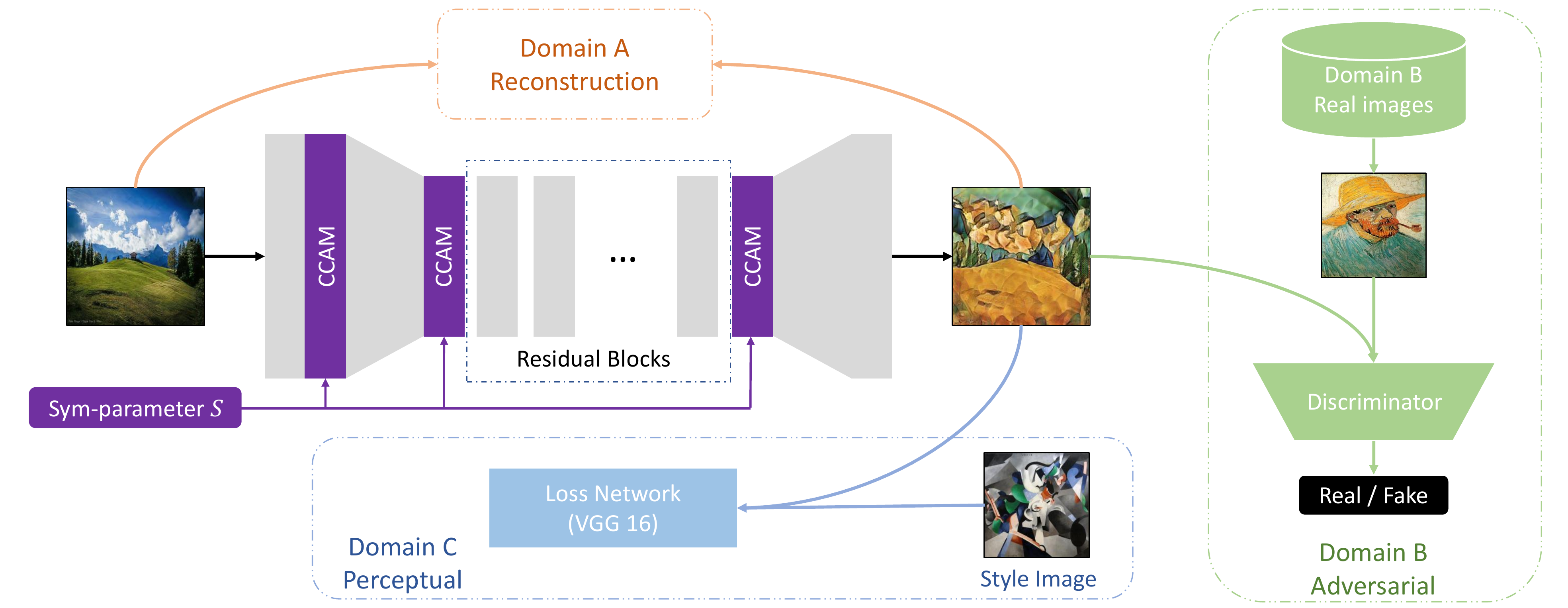}
    \vskip 0.05in
    \caption{\textbf{Overall Structure of SGN for Three Different Losses} 
    This diagram illustrates the case where SGN uses \textit{reconstruction}, \textit{adversarial} and \textit{perceptual} loss as A, B and C domain. For the A, B, and C domains, the SGN uses the weighted sum of the losses with the sym-parameter $S=(s_1, s_2, s_3)$. Therefore, the full objective of the generator is $\mathcal{L}_G=s_1\mathcal{L}_A + s_2\mathcal{L}_B + s_3\mathcal{L}_C$.
}
    \label{fig:SGN}
\end{figure*}

\vskip 0.05in
\noindent\textbf{Training with Dirichlet Distribution}
As mentioned above, a sym-parameter is represented as a vector which has the same number of dimensions as the number of loss functions. The vector's values are randomly selected during training in order to synchronize accordingly with sym-parameterized combinations of loss functions. 
To do so, the sym-parameter values are 
sampled
based on Dirichlet distribution. 
The probability distribution of a $k$-dimensional vector with {the sum of positively valued elements being 1},
can be written as:
{\small
\begin{equation}
p(S) = \frac{1}{B(\alpha)}\prod_{i=1}^{k}s_i^{\alpha_i-1}, \text{ where } B(\alpha) = \frac{\prod_{i=1}^{k}\Gamma(\alpha_i)}{\Gamma(\sum_{i=1}^k\alpha_i)}.
\label{eq:distribution}
\end{equation}
}
Here, $B(\alpha)$ is a normalization constant and $\Gamma(\cdot)$ represents Gamma function. When $k=2$, the distribution {boils down} to Beta distribution.
Using Dirichlet distribution allows the sum of sym-parameter values to be 1 and enables adjusting the distribution with \nj{by changing} the concentration vector $\alpha = (\alpha_1, \cdots, \alpha_k )$.

\subsection{Sym-parameterized Generative Networks}
\label{sec:SGN}
{Using sym-parameters that allow inferences of various mixtures of losses,
we propose sym-parametrized generative networks (SGN) that translate images to \nj{a} mixed-domain. 
Our method is able to either generate images from latent inputs or translate styles with image inputs as long as sym-parameters are inputted along with the inputs and \nj{they} define a linear combination of loss functions. 
Figure \ref{fig:SGN} illustrates the structure of SGN for image-to-image translation.  
Selection of loss functions are not necessarily limited for image generation tasks, and
the following is a representative loss for a generator $G$ with reconstruction loss $\mathcal{L}_{rec}$, adversarial loss $\mathcal{L}_{adv}$ and perceptual loss {$\mathcal{L}_{per}$} weighted by sym-parameters $(s_1, s_2, s_3)$: }
\begin{equation}
\mathcal{L}_G = s_1 \mathcal{L}_{rec} + s_2 \mathcal{L}_{adv} + s_3 \mathcal{L}_{per}.
\label{eq:SGN_lossG}
\end{equation}
Here, each loss function may cope with different 
objective and dataset for more diverse image generations such as using two of adversarial losses one of which for Van Gogh and another for Monet style domain.

While either reconstruction loss or perceptual loss
does not require to train an additional network,
a discriminator must be trained along with an SGN model for the adversarial loss, and a distinct training criterion of the discriminator is needed for SGN.
Because our method generates images based on linearly combined losses with sym-parametrized weights, the weight on the loss of a discriminator must be also set accordingly.
Therefore, discriminator must be trained with the weight assigned on the generator loss in an adversarial manner:
\begin{equation}
\mathcal{L}_D = -s_2 \mathcal{L}_{adv}.
\label{eq:SGN_lossD}
\end{equation}

A trained SGN can translate images with specific sym-parameters, or generate random images from random variables sampled from Dirichlet distribution defined by an input image.

\begin{figure}
    \centering
	\includegraphics[width=1.\linewidth]{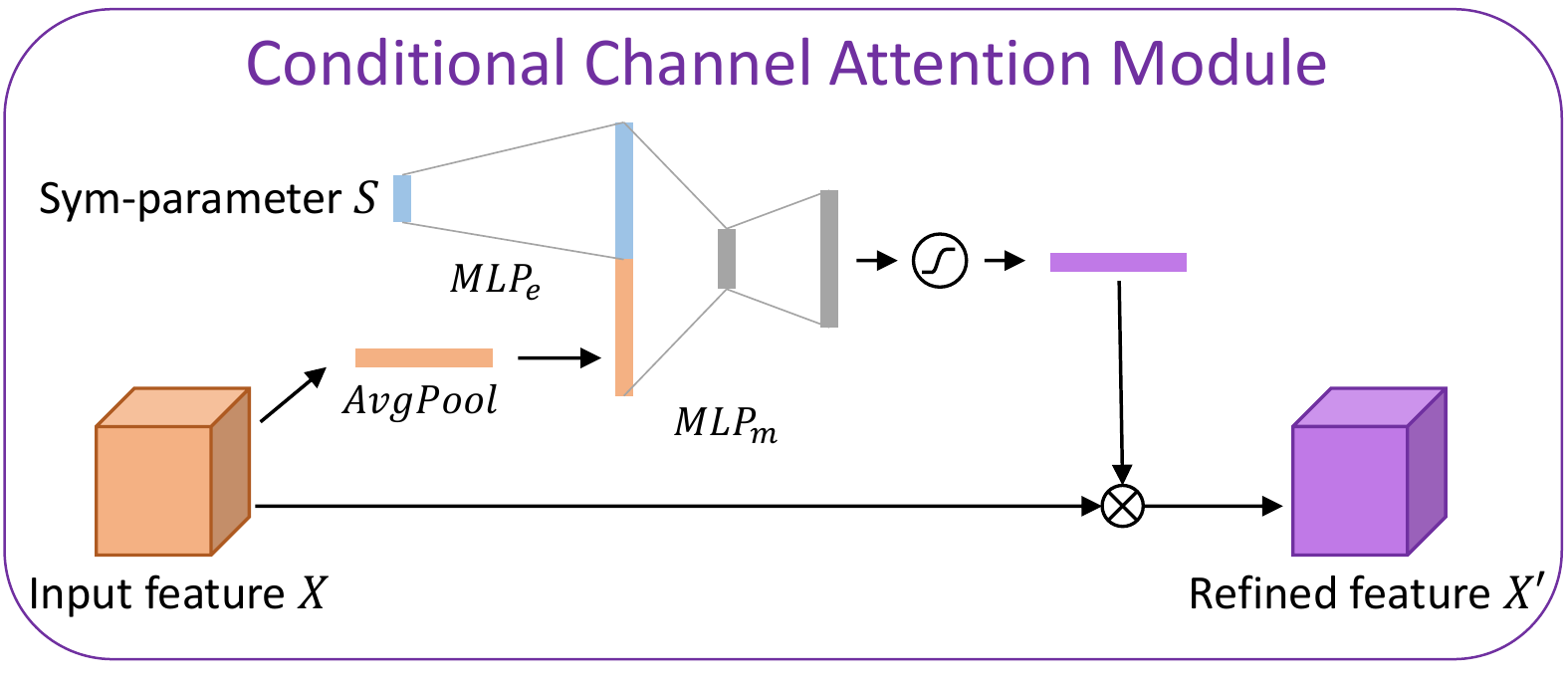}
    \caption{\textbf{Structure of CCAM} CCAM is a lightweight module that takes a feature of previous layer $X$ and a sym-parameter $S$ as an input, generates a map for channel attention through MLP layers, and refines the input feature with this attention map. $\otimes$ denotes a channel-wise multiplication.
}
    \label{fig:CCAM}
\end{figure}

\begin{figure*}[t]
    \centering
	\includegraphics[width=1.\textwidth]{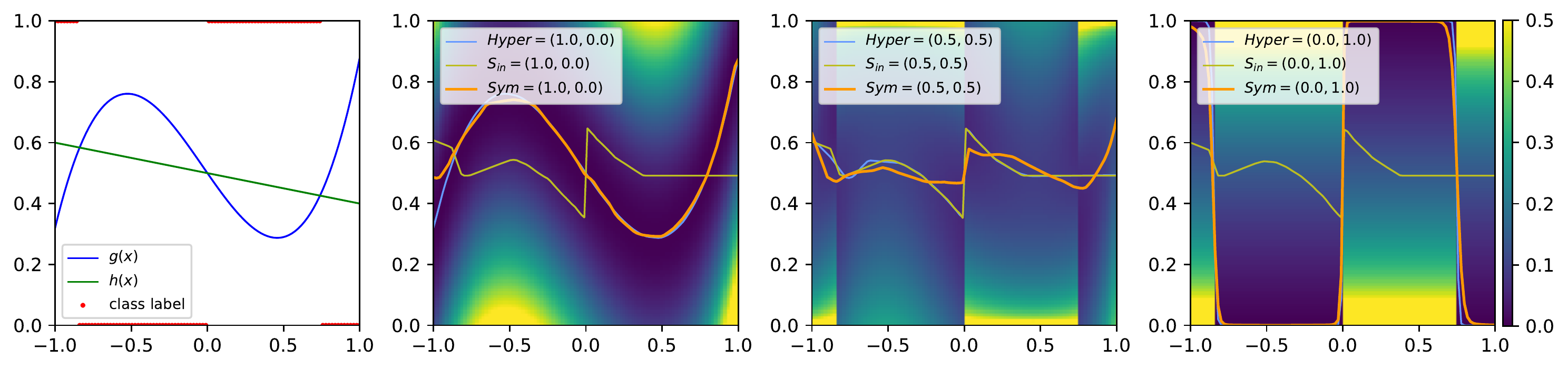}
    \caption{\textbf{Result of 1-D toy problem with sym-parameter.} The left image shows the function $g(x)$, which defines the regression label, and class labels are defined as 1 if $g(x) < h(x)$ or 0 otherwise. \sm{The \hy{color maps}
    on the right \hy{depict}
    the calculated loss value\hy{s} for $L(f, S)$ and the plot \textit{Sym} is the actual output of the trained $f(x, S)$ on the given $S$. \textit{Hyper} denotes the output of multiple hyper-parameter models for each weights. $S_{in}$ \nj{uses the} sym-parameter as \nj{an} input, but \nj{uses} fixed weight (0.5, 0.5) for each loss. This result shows that our method follows the weighted loss according to the sym-parameter $S$. }}
    \label{fig:toy}
\end{figure*}

\subsubsection{Conditional Channel Attention Module}
SGN takes a continuous valued sym-parameter {vector} along with inputs and generates images {reflecting characteristics} of various parts of a mixed-domain, which covers wider range of target distribution than multi-domain models do with discrete conditions.
Since domain injection used in conventional conditional generators is empirically shown to be inadequate for our purpose, we propose another injection method for sym-parameterized conditionings, named Conditional Channel Attetion Module (CCAM). 
Inspired by SENet \cite{hu2017senet}, CCAM is a channel attention model that selectively gates feature channels based on sigmoid attentiveness. 
CCAM also allows SGN to have {a} fully convolutional structure and manage various spatial sizes.
CCAM's structural details are depicted in Figure \ref{fig:CCAM}, and the module can be written as:
{\small
\begin{equation}
CCAM(X,S) = X \cdot \sigma(MLP_m([MLP_e(S), AvgPool(X)])),
\label{eq:CCAM}
\end{equation}
}
where {$X \in \mathbb{R}^{H \times W \times C}$} represents feature maps outputted from a preceding layer {and $[\cdot, \cdot]$ denotes the concatenation operation.} The features are shrunk to $1\times 1 \times C$ through an average pooling, and the sym-parameters are represented in the same size as the pooled features through $MLP_e$.
After $MLP_m$ creates attention maps based on the concatenation of input features and sym-parameters, 
a sigmoid function $\sigma$ activates to produce the channel attention map, $M \in \mathbb{R}^{1 \times 1 \times C}$.
The output features of CCAM is then created by channel-wise multiplication of the channel attention map $M$ and the original feature map $X$.
For such a continuous conditioning case, CCAM allows superior efficiency of image generations over channel-wise inclusion of domain information along with RGB channels \cite{choi2017stargan} or domain code injection through central biasing normalization \cite{yu2018singlegan}.

\section{Implementation of SGN}
\label{sec:implementation}
\noindent\textbf{Network Architecture}
The base architecture of SGN {has} adopted the autoencoder layers from \cite{choi2017stargan,johnson2016perceptual,zhu2017unpaired}, which consists of two downsampling convolution layers, two upsampling transposed convolution layers with strides of two and nine residual blocks
in between them. 
In this architecture, CCAM modules are inserted for conditioning at three positions, before the downsampling convolutions, after the downsampling convolutions and lastly, before the upsampling convolutions. 
For $\mathcal{L}_{adv}$, we use PatchGAN discriminator, introduced by Isole \textit{et al.} \cite{isola2017image} where discriminator is applied at each image patch separately. 
The choice of feed-forward CNN for $\mathcal{L}_{per}$ is VGG16 with consensus on the work of Johnson \textit{et al.} \cite{johnson2016perceptual}. 
More details of the architecture and training will be handled on supplementary materials.

\section{Experiments}
Since, to the best of our knowledge, our method is the first approach that allows neural networks to dynamically adjust the balance among losses at inference time, there are no baselines for direct comparisons.
We perform experiments for qualitative and quantitative evaluations on how well the sym-parameter works. And to focus on fair investigations based on the results,
we have preserved the models of the existing methods and their individual losses as reported except for few minor changes.
In this section, we first investigate the behavior of sym-parameters based on different loss functions through a 1-D toy problem. 
We then experiment an SGN on image translations to mixed-domains.
And lastly, CCAM's role within the SGN is reviewed.

\begin{table}[t]
\resizebox{1.\linewidth}{!}{
\renewcommand{\arraystretch}{1.2}
\hskip -0.05in
\begin{tabular}{c|lll|lll}
\hline
Weights & \multicolumn{3}{c|}{\textbf{Sym-parameter model}} & \multicolumn{3}{c}{\textbf{Hyper-parameter models}} \\
($\mathcal{L}_r$, $\mathcal{L}_c$) & \multicolumn{1}{c}{$\mathcal{L}$} & \multicolumn{1}{c}{$\mathcal{L}_r$} & \multicolumn{1}{c|}{$\mathcal{L}_c$} & \multicolumn{1}{c}{$\mathcal{L}$} & \multicolumn{1}{c}{$\mathcal{L}_r$} & \multicolumn{1}{c}{$\mathcal{L}_c$} \\
\hline
(1.00, 0.00) & 0.0001 & \textbf{0.0001} & 0.2148 & 0.0000 & \textbf{0.0000} & 0.2167 \\
(0.75, 0.25) & 0.0483 & 0.0063 & 0.1743 & 0.0481 & 0.0062 & 0.1737 \\
(0.50, 0.50) & 0.0824 & 0.0347 & 0.1302 & 0.0815 & 0.0353 & 0.1277 \\
(0.25, 0.75) & 0.0878 & 0.1674 & 0.0613 & 0.0846 & 0.1597 & 0.0592 \\
(0.00, 1.00) & 0.0062 & 0.4113 & \textbf{0.0062} & 0.0013 & 0.4254 & \textbf{0.0013} \\
\hline
\end{tabular}
}
\hskip -0.05in
\vskip 0.1in
\caption{
\textbf{Comparison against hyper-parameter models on 1D toy problem. }
The single model with sym-parameter has a similar loss value to those from the hyper-parameter models learned separately. $\mathcal{L}_r$ and $\mathcal{L}_c$ denote the regression loss and the classification loss, respectively. $\mathcal{L}$ is weighted loss of $\mathcal{L}_r$ and $\mathcal{L}_c$ with the given weight parameters.
}
\label{table:sym-loss}
\end{table}

\begin{figure*}
    \centering
	\includegraphics[width=0.98\textwidth]{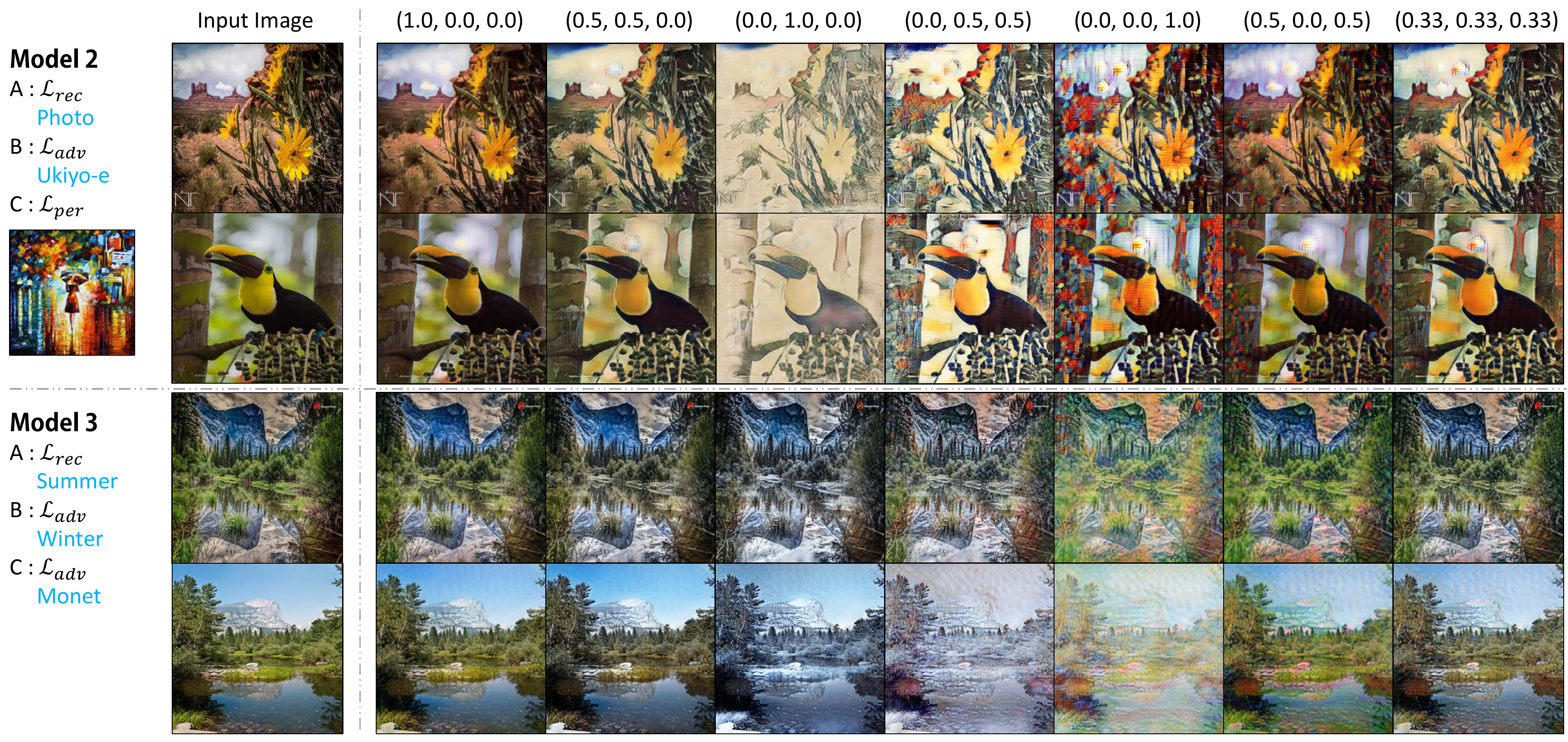}
    \caption{\textbf{Image Translation Results.} These images are translated from the test dataset with SGN model 2 and 3 according to the given sym-parameters $S$. The numbers in parentheses are the sym-parameters for the A, B, and C domains, respectively. }
    \label{fig:SGN_result2}
\end{figure*}

\subsection{Toy Example: Regression and Classification with Single Network}
\label{sec:toy}
\sm{In order to understand sym-parameters, we \hy{have} designed an 1-D toy problem that we can calculate exact loss and visualize it. This allows us to confirm that our method can minimize multiple set of weights of multiple losses. And we can compare with hyper-parameter \nj{models} whether a single sym-parameter model can replace various hyper-parameter models.}
We have defined a polynomial function $g(x)$ that takes one dimensional vector $x \in [-1,1]$ and outputs $y_r$.
Also, with $g(x)$ and a linear function $h(x)$, $y_c$ is represented as a binary class label of 1 if $g(x)<h(x)$ or 0 otherwise. 
We have created a dataset consisting of ($x, y_r, y_c$) tuples. 
A sym-parametrized \nj{MLP} network $f(x,S)$ concisely structured with three hidden layers is trained with the dataset to perform regression and classification in terms of $y_r$ and $y_c$, respectively.
The total loss is defined as $\mathcal{L}(f,S) = s_1\mathcal{L}_r + s_2\mathcal{L}_c$ with a sym-parameter $S=(s_1,s_2)$.

\begin{figure*}
    \centering
    \includegraphics[width=0.98\textwidth]{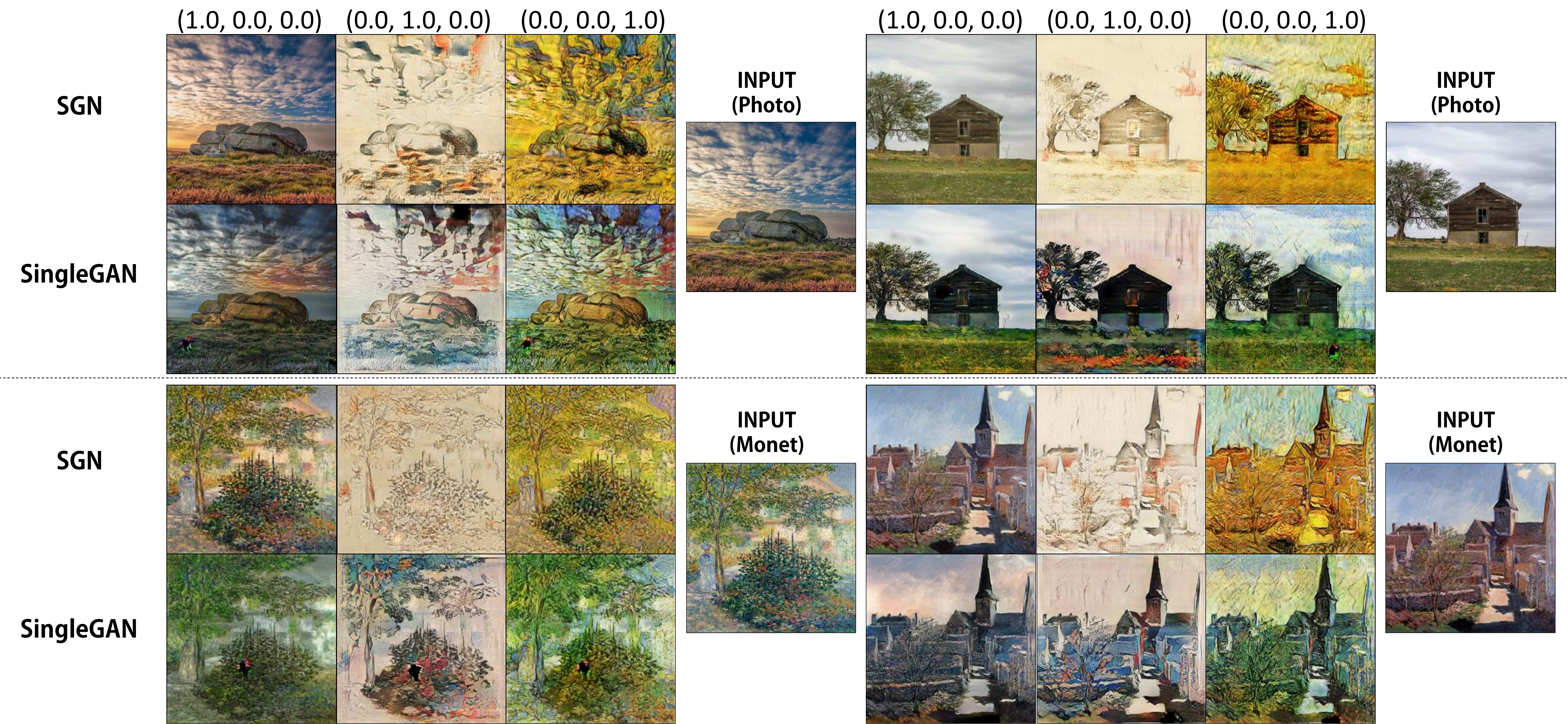}
    \caption{\smn{\textbf{Comparison with \knj{Style Transfer Model using Multiple Datasets}} SingleGAN can generate images with multiple domains, but the output relies on trained datasets. Therefore it cannot reconstruct image even if it transfers the pictures of the same domain. But SGN can reconstruct new domain images like Monet because it is trained with reconstruction losses for sym-parameter S(1,0,0).}
    }
    \label{fig:one2many}
\end{figure*}

\begin{table}[t]
\resizebox{1.\linewidth}{!}{
{\renewcommand{\arraystretch}{1.2}
\hskip -0.05in
\begin{tabular}{l|lll}
\hline
        & \multicolumn{1}{c}{A} & \multicolumn{1}{c}{B} & \multicolumn{1}{c}{C}     \\
\hline
Model 1 & $\mathcal{L}_{rec}$, Photo      & $\mathcal{L}_{adv}$, Van Gogh   & $\mathcal{L}_{per}$, Udnie         \\
Model 2 & $\mathcal{L}_{rec}$, Photo      & $\mathcal{L}_{adv}$, Ukiyo-e    & $\mathcal{L}_{per}$, Rain \\
Model 3 & $\mathcal{L}_{rec}$, Summer     & $\mathcal{L}_{adv}$, Winter     & $\mathcal{L}_{adv}$, Monet         \\
\hline
\end{tabular}
\hskip -0.05in
}
}
\vskip 0.1in
\caption{\textbf{The configuration of models.} We configure A, B and C domains for the three models used in the SGN experiments. Each domain is defined as a combination of loss and data.}
\label{table:models}
\end{table}

Figure \ref{fig:toy} shows the results of the experiment.
Color-maps on the right sub-figures show computed loss values when weight for each losses are $(1,0), (0.5,0.5)$ and $(0,1)$, respectively. And the plots in each sub-figure denote the results of sym-parameter model ($Sym$), hyper-parameter models ($Hyper$) and \nj{an} additional ablation ($S_{in}$).
The figures clearly show \nj{the} sym-parametrized model \nj{$f(x,S)$} properly minimizes weighted losses according to \nj{the sym-parameter value}. 
And the results are similar to the multiple hyper-parameter models, which are trained individually. 
\sw{Model $S_{in}$ \nj{uses the} sym-parameter as an input}, but the weights of each losses are \nj{fixed} to $(0.5,0.5)$ \nj{at training.} As the illustrated result\nj{s} show, outputs follow the case $H = (0.5,0.5)$ \nj{regardless of the input $S$ because the model $S_{in}$ did not learn to change the loss function.} 
Also, the same sym-parametrized model is quantitatively compared against various hyper-parameter models that are separately trained.
As can be seen in Table~\ref{table:sym-loss}, the loss values of the models separately trained with hyper-parameterized weights do not differ much from the values of the single model trained with sym-parametrized weights.
This implies that a single training of a model using sym-parameters may replace the models trained in multiple cases of hyper-parameters.

\subsection{Image Translation to Mixed-Domain}
We have set up three SGN models for experiments on image translations to mixed-domain. 
With the use of sym-parameters, SGN extends the concept of multi-domain to loss functions, 
and each domain is thus defined with combinations of loss functions and data. 
For our experiments, the datasets used in \cite{johnson2016perceptual,zhu2017unpaired} are combined with loss criteria of $\mathcal{L}_{rec}$, $\mathcal{L}_{adv}$ and $\mathcal{L}_{per}$ from Section~\ref{sec:implementation} to train each model as presented in Table~\ref{table:models}.

\begin{table}[t]
\resizebox{1.\linewidth}{!}{
\centering
\renewcommand{\arraystretch}{1.2}
\renewcommand{\tabcolsep}{0.17cm}
\hskip -0.15in
\begin{tabular}{c|cccc|cccc}
\hline
Weights & \multicolumn{4}{c|}{\textbf{Single sym-parameter model}} & \multicolumn{4}{c}{\textbf{Hyper-parameter models}} \\
{\small$(\mathcal{L}_{rec},\mathcal{L}_{adv}, \mathcal{L}_{per})$} & $\mathcal{L}_G$ & $\mathcal{L}_{rec}$ & $\mathcal{L}_{adv}$ & $\mathcal{L}_{per}$ & $\mathcal{L}_G$ & $\mathcal{L}_{rec}$ & $\mathcal{L}_{adv}$ & $\mathcal{L}_{per}$\\
\hline
(1.0, 0.0, 0.0) & 0.164 & 0.164 & 0.829 & 0.209 &0.159&0.159&2.540&0.252\\
(0.5, 0.5, 0.0) & 0.305 & 0.285 & 0.323 & 0.253 &0.407&0.423&0.391&0.270\\
(0.0, 1.0, 0.0) & 0.211 & 0.756 & 0.211 & 0.311 &0.308&1.008&0.308&0.276\\
(0.0, 0.5, 0.5) & 0.171 & 0.718 & 0.256 & 0.085 &0.205&0.924&0.247&0.163\\
(0.0, 0.0, 1.0) & 0.040 & 0.443 & 0.525 & 0.040 &0.015&1.039&1.375&0.015\\
(0.5, 0.0, 0.5) & 0.139 & 0.201 & 0.889  &0.077  &0.117&0.184&3.054&0.050\\
\hline
\end{tabular}
}
\label{table:loss_SGN}
\caption{\textbf{Comparisons with six hyper-parameter models in image-to-image translation domain.} The weighted loss, $\mathcal{L}_G$, of the single SGN has a similar test error to the hyper-parameter models trained separately for each set of weights. This experiment uses the setting of \textit{Model 1} in the Table 2.}
\end{table}

\vskip 0.02in
\noindent\textbf{Qualitative Result }
Figure \ref{fig:main} illustrates translation results of Model 1 with various sym-parameters. 
Not only inter-translations among A, B and C domains are well generated, 
but the model also produces mixed generations with weighted characteristics of candidate domains.  
Especially, the successful image translations between domain B with adversarial loss and domain C with perceptual loss should be noticed; the generated images are influenced more by colors and styles of Van Gogh than by the test image input.
As the numbers in the parentheses represent sym-parameter values, correspondingly styled images are well produced when mixed with 3 types of domains.
Lastly, generated image with $S=(0.0, 1.5, 0.0)$ is an extrapolated result with a sym-parameter that is set outside of the trained range, and Van Gogh's style is still bolstered accordingly. 

Figure~\ref{fig:SGN_result2} shows image translations yielded by two other models. 
While an SGN can be trained with different loss functions and datasets as done for Model 1 and 2,
it can also be trained with domains defined with two GAN losses for different datasets.
As it can be seen in the figure, Model 3 is able to translate a summer Yosemite image not only to a winter Yosemite image but also to a winter image with Monet style with $S=(0.0,0.5,0.5)$.
This result well depicts how weighted optima of multiple objective functions can be expressed. 
More test results of each model are provided in the supplementary material.

\vskip 0.02in
\noindent\textbf{Quantitative Result }
It is generally known to be difficult to define quantitatively evaluating metrics for generative models.
Nonetheless, it is logical to quantitatively examine by checking loss values for different $S$ values since our aim is to find optimal of variously weighted sum of losses.
If trained sufficiently in terms of (\ref{eq:sym-param}) and (\ref{eq:SGN_lossG}), 
the corresponding domain's loss should be minimized when a domain's weight is 1, 
and when the weight is mid-valued as 0.5, resultant loss of the domain should be valued in between the values when the weight is 1 and 0. 
\sm{\nj{Furthermore, we can compare these values with} multiple hyper-parameter models that are individually trained.}

\begin{figure*}
    \centering
    \includegraphics[width=0.98\textwidth]{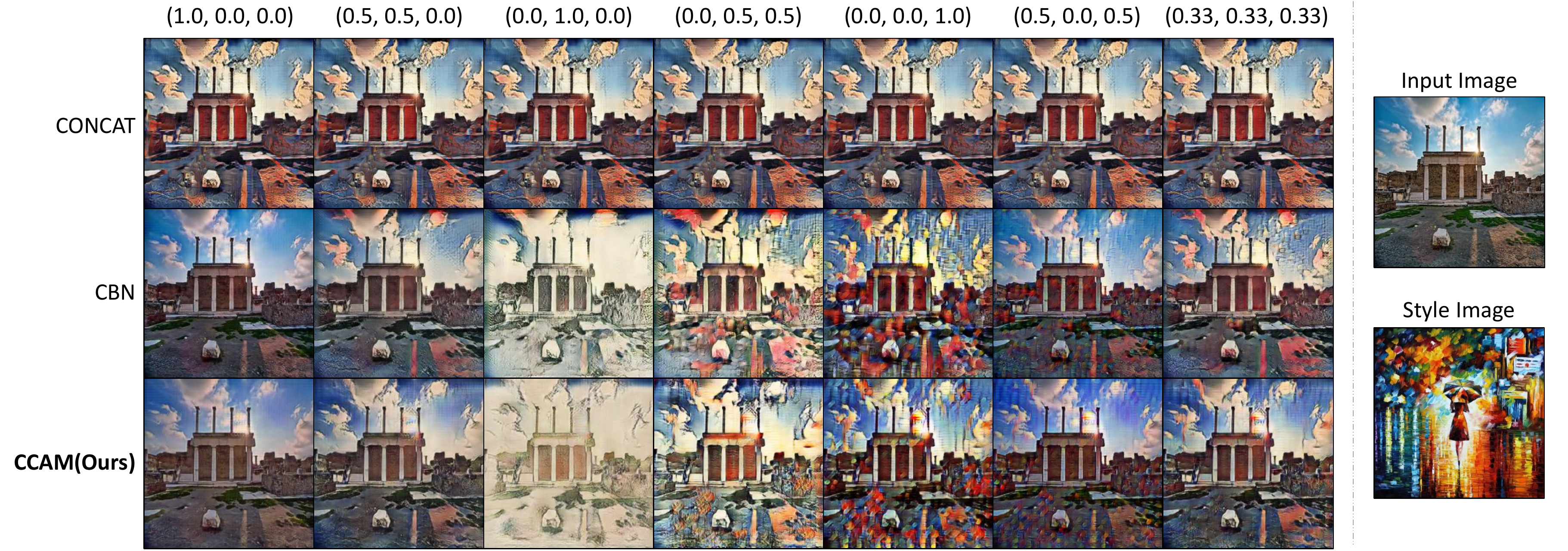}
    \caption{\textbf{Results based on different injection methods of sym-parameters.} 
    All settings except for the sym-parameter injection method are equivalently set up as Model 2 (A: $\mathcal{L}_{rec}$, B: $\mathcal{L}_{adv}$, Ukiyo-e, C: $\mathcal{L}_{per}$). 
    }
    \label{fig:injection}
\end{figure*}

Loss values of trained Model 1 are measured with a test dataset and enumerated in Table~\ref{table:loss_SGN}. 
The numbers in the table represent averaged loss values of each loss function the generator uses. 
In the table, each domain yields its minimum loss value when weighted with 1. And these are similar to the multiple models trained using \nj{a corresponding hyper-parameter}. 
The largest difference is that the hyper-parameter models have a high loss value when the weight of a specific loss is zero at the time of testing, such as $ S = (0, 0.5, 0.5) $, while the SGN has a relatively low value. This is because it is advantageous for the SGN to learn that the loss is minimized for similar $S$ values such as $ S = (0.1, 0.45, 0.45) $.

\vskip 0.02in
\noindent\textbf{Multiple Datasets vs Multiple Losses} 
We have performed a comparison with SingleGAN~\cite{yu2018singlegan}, which can handle multiple datasets, to clearly show the difference of our multi-loss approach from the conventional methods. For fair comparison, SingleGAN has been trained using Photo as input with three \knj{output domains:} Photo, Ukiyo-e and Van Gogh. SGN used \nj{the} reconstruction loss for Photo, GAN loss\knj{es} for Ukiyo-e and Van Gogh. As shown in Figure \ref{fig:one2many}, SingleGan performs style transfer, not reconstruction, for the condition (1,0,0). These results mean that the input images are transferred to Photo style. This can be seen more clearly by using Monet images as input, which are not used for training.
In contrast, SGN performs reconstruction for $\mathcal{S} = (1, 0, 0)$ according to the definition of the trained loss, and also operates to reconstruct Monet images that are not used in the training phase at all. This result indicates that SGN behaves according to the loss characteristics learned as we intended.

\vskip 0.02in
\noindent\textbf{Continuous Translation }
Since the sym-parameter is represented as a $k$-dimensional continuous vector, continuous inter-domain translations are achieved during mixed-domain image translations. 
Such characteristics of SGN allows seamless mixed-domain transitions among candidate domains through the video\footnote{\url{https://youtu.be/i1XsGEUpfrs}}. 

\subsection{Conditional Channel Attention Module}
We perform sym-parameter injections into SGN through CCAM since conventional condition injecting methods are empirically shown inefficient for our method. 
We have thus experimentally compared the performance results among different cases of condition injections, which is depicted in Figure \ref{fig:injection}. 
The injection method labeled CONCAT represents the method of depth-wise concatenations with a latent coded vector repeated to be in the same spatial size as features. 
Central biasing normalization (CBN) \cite{yu2018multi} is also experimented for comparisons. 
We have adjusted the instance normalization of the down-sampling layers and the residual blocks of the generator network as similarly used in \cite{yu2018singlegan}. 

CONCAT method produces outputs that are alike despite of various combinations of sym-parameters and fails to generate images in mixed-domain correspondingly.
This method works promisingly for discrete condition injections as reported in \cite{choi2017stargan,lample2017fader,zhu2017toward}
but struggles in the SGN using continuous conditions. 
This phenomenon is reasonable considering that the normalization within the generator perturbs the values of the sym-parameter and differences in the values are susceptible. 
CBN method performs better than CONCAT, generating comparably more various images with given sym-parameters. 
Yet, its generations are rather biased to one domain than a mixed-domain targeted by sym-parameters, and also follow with some artifacts. 
Since CBN uses bias and thus is hard to exclude the possibility on influence of a particular channel, additional interferences among candidate domains may occur. 
Among the experimented condition injecting methods, our proposed CCAM is more suitable for the sym-parameter and SGN.

\section{Conclusion}
In this paper, we propose a sym-parameter that can extend the concept of domain from data to loss and adjust the weight among multiple domains at inference. We then introduce SGN, which is a novel network that can translate image into mixed-domain as well as each domain using this sym-parameter. It is hard to say which method is proper and optimal when it comes to making a valid result for a mixed-domain without ground truth. However, if optimizing to the weighted objective of each domain is one of the effective methods for this purpose, SGN performs well to translate image to a target mixed-domain as shown in the experiments. 
We expect that the research will be extended to apply sym-parameter to more diverse domains, and to find more effective models for sym-parameter.

\vskip 0.05in
{\noindent\textbf{Acknowledgement}\\}
This work was supported by Next-Generation Information Computing Development Program through the National Research Foundation of Korea (NRF-2017M3C4A7077582). 

{\small
\bibliographystyle{ieee_fullname}
\bibliography{egbib}
}
\clearpage

\onecolumn
\maketitle

\noindent\Large{\textbf{Supplementary Material}}
\setcounter{section}{0}
\normalsize
\renewcommand\thesection{\Alph{section}}
\section{Experiment Settings}
\subsection{1D Toy Example}
The polynomial $g(x)$ and the linear function $h(x)$ are defined such that the output is between 0 and 1 when the range of $x$ is -1 to 1.
\begin{equation}
g(x) = x(x-0.8)(x+0.9) + 0.5
\label{eq:toy_g}
\end{equation}
\begin{equation}
h(x) = {-0.1x} + 0.5 
\label{eq:toy_h}
\end{equation}
The sym-parametrized network $f(x,S)$ is an MLP network that has three hidden layers with the size of 64 and ReLU activations. It takes a tuple of $(x, S)$ as an input and outputs a single value for either classification or regression. The mean squared error function is used for regression loss $\mathcal{L}_r$, and binary cross entropy is used for the classification loss $\mathcal{L}_c$.  $\mathcal{L}_c$ is scaled down to 20\% to balance the losses. In the training phase, the sym-parameter $S$ is sampled from Dirichlet distribution with $\alpha$ valued (0.5, 0.5). We use ADAM with a batch size of 16, and learning rates of
0.01 during \nj{the} first 200 epochs, 0.001 during \nj{the} next 200 epochs and 0.0001 during \nj{the} last
100 epochs.

\subsection{Sym-Parameterized Generative Network}
\smn{For three loss terms, $\mathcal{L}_{rec}$, $\mathcal{L}_{adv}$ and $\mathcal{L}_{per}$, we use ${L}_{1}$ norm for the reconstruction loss $\mathcal{L}_{rec}$, and have applied a technique of LSGAN \cite{mao2017least} for the adversarial loss $\mathcal{L}_{adv}$ in which an MSE (mean squared error) loss is used.
Additionally, the identity loss introduced by Taigman \textit{et al.} \cite{taigman2016unsupervised} and used in CycleGAN \cite{zhu2017unpaired} is used in $\mathcal{L}_{adv}$ for regularizing the generator, combined with the LSGAN loss. 
The implementation of the perceptual loss $\mathcal{L}_{per}$ follow the implementation of Johnson \textit{et al.} \cite{johnson2016perceptual}, adopting both feature reconstruction loss and style reconstruction loss. 
At the training phase, all three loss terms were weighted by numbers sampled from Dirichlet distribution with all the $\alpha$ valued 0.5.}

The architecture details of an SGN generator is provided in Table \ref{table:setting_G} \nj{of this supplementary}. 
The discriminator is equivalently set as CycleGAN~\cite{zhu2017unpaired}. 
To regulate the imbalance between the losses, $\mathcal{L}_{rec}$ and $\mathcal{L}_{adv}$ are respectively weighted with 2 and 1. And for $\mathcal{L}_{adv}$, GAN loss and the identity loss are respectively weighted \nj{by} 1 and 5.
For the perceptual loss $\mathcal{L}_{per}$, we have used \nj{the} pretrained Pytorch model of VGG16 \cite{DBLP:journals/corr/SimonyanZ14a} without batch-normalization layers. 
$\mathcal{L}_{per}$ is composed of a content loss and a style loss, computed at first 4 blocks of VGG16, which means it uses output features from Conv1-2, Conv2-2, Conv3-3, Conv4-3, and do not use the fifth block, following the work of Johnson \textit{et al.} \cite{johnson2016perceptual}. 
The content loss is weighted with 0.001 and the style losses at 4 layers \nj{are} weighted as (0.1, 1.0, 10, 5.0)$\times 200$ for the Model 1,  (0.1, 1.0, 10, 5.0)$\times 100$ for the other models. We use the ADAM optimizer with a batch size of 4. Then Model 1 and Model 2 are trained for 20 epochs and Model 3 is trained for 60 epochs. We keep the learning rate of 0.0002 for the first half of epochs and linearly decay \nj{it} to zero for the remaining epochs. \smn{Source code and pre-trained networks are available in \url{https://github.com/TimeLighter/pytorch-sym-parameter}}.

\begin{table}
\centering
\resizebox{1.\linewidth}{!}{
{\renewcommand{\arraystretch}{1.2}
\hskip -0.05in
\begin{tabular}{c|c|c|c}
\hline
Layer Configuration & Input Dimension & Layer Information & Output Dimension\\
\hline
Input convolution & (${h}$, ${w}$, 3) & Convolution (K:7x7, S:1, P:3), IN, ReLU & (${h}$, ${w}$, 64) \\
\cline{1-4}
\multicolumn{4}{c}{CCAM (Reduction Rate $r$: 4) }\\
\cline{1-4}
\multirow{2}{*}{Down-sampling} & (${h}$, ${w}$, 64) & Convolution (K:3x3, S:2, P:1), IN, ReLU & ($\frac{h}{2}$, $\frac{w}{2}$, 128)\\
& ($\frac{h}{2}$, $\frac{w}{2}$, 128) & Convolution (K:3x3, S:2, P:1), IN, ReLU  &  ($\frac{h}{4}$, $\frac{w}{4}$, 256) \\
\cline{1-4}
\multicolumn{4}{c}{CCAM (Reduction rate $r$: 4)}\\
\cline{1-4}
\multirow{9}{*}{Residual Blocks} 
& ($\frac{h}{4}$, $\frac{w}{4}$, 256) & Convolution (K:3x3, S:1, P:1), IN, ReLU  & ($\frac{h}{4}$, $\frac{w}{4}$, 256) \\
& ($\frac{h}{4}$, $\frac{w}{4}$, 256) & Convolution (K:3x3, S:1, P:1), IN, ReLU  & ($\frac{h}{4}$, $\frac{w}{4}$, 256) \\
& ($\frac{h}{4}$, $\frac{w}{4}$, 256) & Convolution (K:3x3, S:1, P:1), IN, ReLU  & ($\frac{h}{4}$, $\frac{w}{4}$, 256) \\
& ($\frac{h}{4}$, $\frac{w}{4}$, 256) & Convolution (K:3x3, S:1, P:1), IN, ReLU  & ($\frac{h}{4}$, $\frac{w}{4}$, 256) \\
& ($\frac{h}{4}$, $\frac{w}{4}$, 256) & Convolution (K:3x3, S:1, P:1), IN, ReLU  & ($\frac{h}{4}$, $\frac{w}{4}$, 256) \\
& ($\frac{h}{4}$, $\frac{w}{4}$, 256) & Convolution (K:3x3, S:1, P:1), IN, ReLU  & ($\frac{h}{4}$, $\frac{w}{4}$, 256) \\
& ($\frac{h}{4}$, $\frac{w}{4}$, 256) & Convolution (K:3x3, S:1, P:1), IN, ReLU  & ($\frac{h}{4}$, $\frac{w}{4}$, 256) \\
& ($\frac{h}{4}$, $\frac{w}{4}$, 256) & Convolution (K:3x3, S:1, P:1), IN, ReLU  & ($\frac{h}{4}$, $\frac{w}{4}$, 256) \\
& ($\frac{h}{4}$, $\frac{w}{4}$, 256) & Convolution (K:3x3, S:1, P:1), IN, ReLU  & ($\frac{h}{4}$, $\frac{w}{4}$, 256) \\
\cline{1-4}
\multicolumn{4}{c}{CCAM (Reduction Rate $r$: 4)}\\
\cline{1-4}
\multirow{2}{*}{Up-sampling} 
& ($\frac{h}{4}$, $\frac{w}{4}$, 256) & Transposed Convolution (K:3x3, S:2, P:1), IN, ReLU & ($\frac{h}{2}$, $\frac{w}{2}$, 128) \\
& ($\frac{h}{2}$, $\frac{w}{2}$, 128) & Transposed Convolution (K:3x3, S:2, P:1), IN, ReLU & (${h}$, ${w}$, 64) \\
\cline{1-4}
Output Convolution & (${h}$, ${w}$, 64) & Convolution (K:7x7, S:7, P:3), Tanh & (${h}$, ${w}$, 3) \\
\hline
\end{tabular}
\hskip -0.05in
}
}
\vskip 0.1in
\caption{
\textbf{Generator network architecture.} In the Layer Information column, K: size of the filter, S: stride size, P: padding size, IN: Instance Normalization.
CCAM has reduction rate $r$ to reduce the amount of computation like SENet~\cite{hu2017senet}.
}
\label{table:setting_G}
\end{table}

\clearpage
\section{Additional Experimental Results}
\subsection{Tradeoff with Network Size}
\begin{figure}[h]
    \centering
	\includegraphics[width=0.6\linewidth]{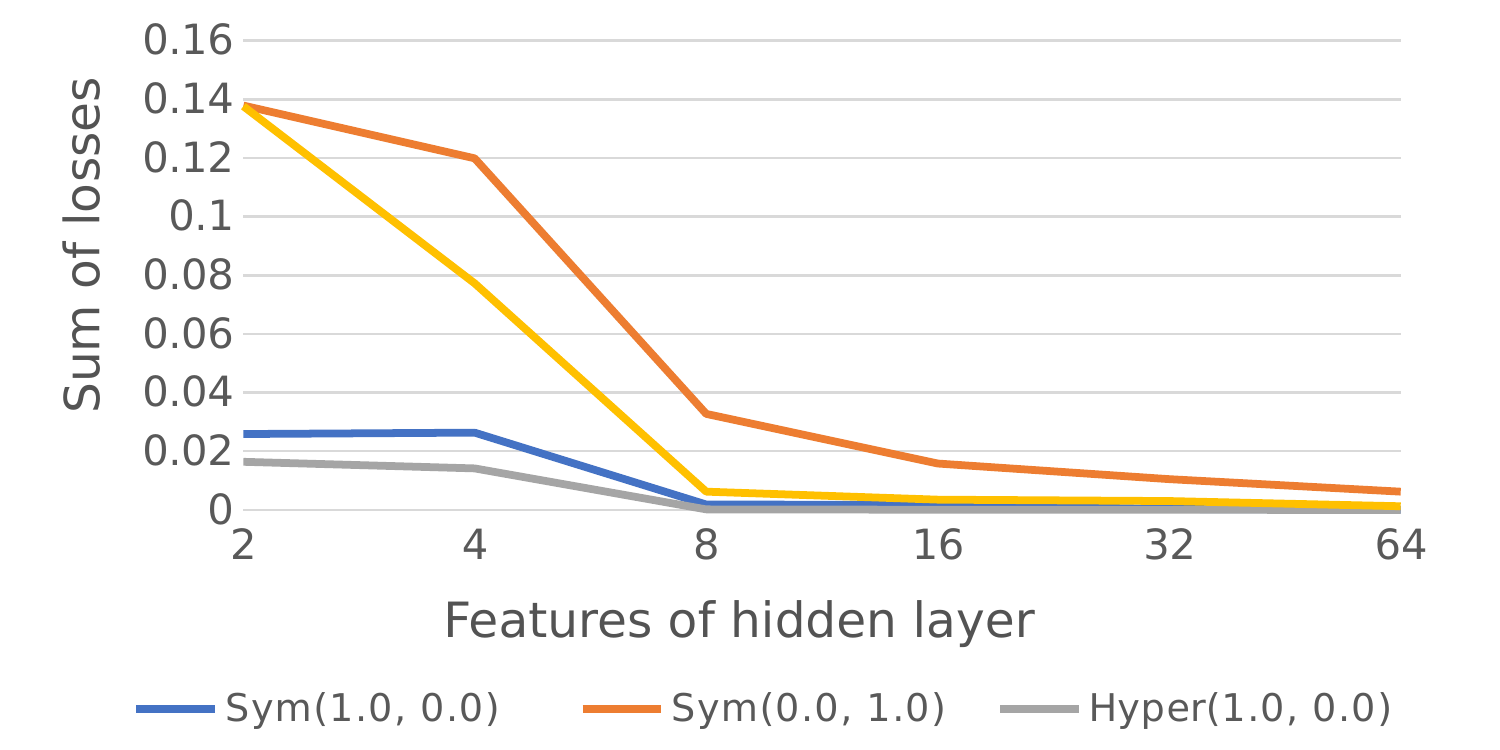}
    \caption{\textbf{Effect of Network Size} }
    \label{fig:size}
\end{figure}

Using sym-parameter does not always have tradeoffs in performance, but it is affected by the capacity of the model. Figure \ref{fig:size} shows the losses when changing the number of filters in the 1D Toy problem. As the size of the model decreases, the loss is getting larger than the hyper-parameter models. However, considering that a single sym-parameter model can be used as multiple hyper-parameter models with a specific range of weights, the difference is not significant.

\subsection{Continuous Translation}
The video (\url{https://youtu.be/i1XsGEUpfrs}) is created by extracting images from the original and translating it through SGN.
\textit{ORIGINAL VIDEO} and \textit{VIDEO WITH SGN} represent original video and SGN translated video, respectively. We use SGN models trained in the paper and translated images changing only the sym-parameters. 
The color bar above \textit{VIDEO WITH SGN} represents the sym-parameter values for each domain. If the entire bar is red, then the sym-parameter $S$ is (1,0,0). Since all the images are translated through SGN, reconstructed images may differ in some colors and appearance from the original ones.

\subsection{More Results of CCAM}
\begin{figure}
\centering
\subfigure{\includegraphics[width=0.5\linewidth,height=3.5in]{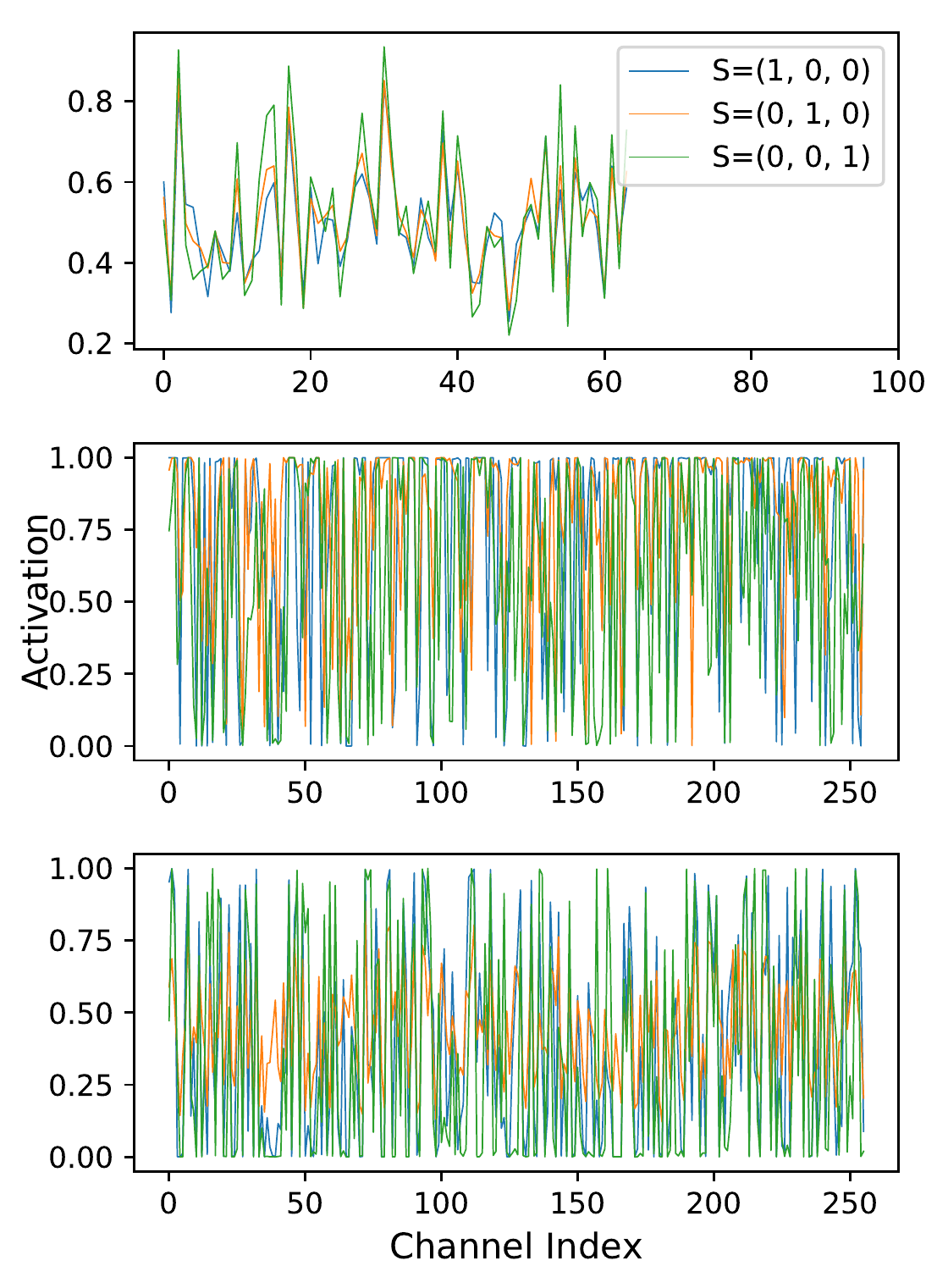}}
\subfigure{\includegraphics[width=0.15\linewidth]{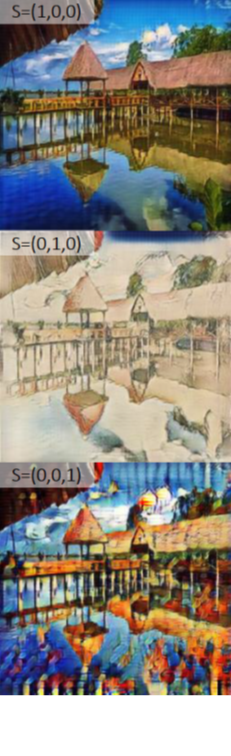}}
    \caption{\textbf{Channel activation of CCAMs.} Each plot shows the channel activation results of the three CCAMs used in the SGN. The lower plot corresponds to the CCAM of deeper layer. This result indicates that the degree of activation is different for each channel when the sym-parameter is different for the same image.}
    \label{fig:ch_act}
\end{figure}

\begin{figure}
    \centering
	\includegraphics[width=1.\linewidth]{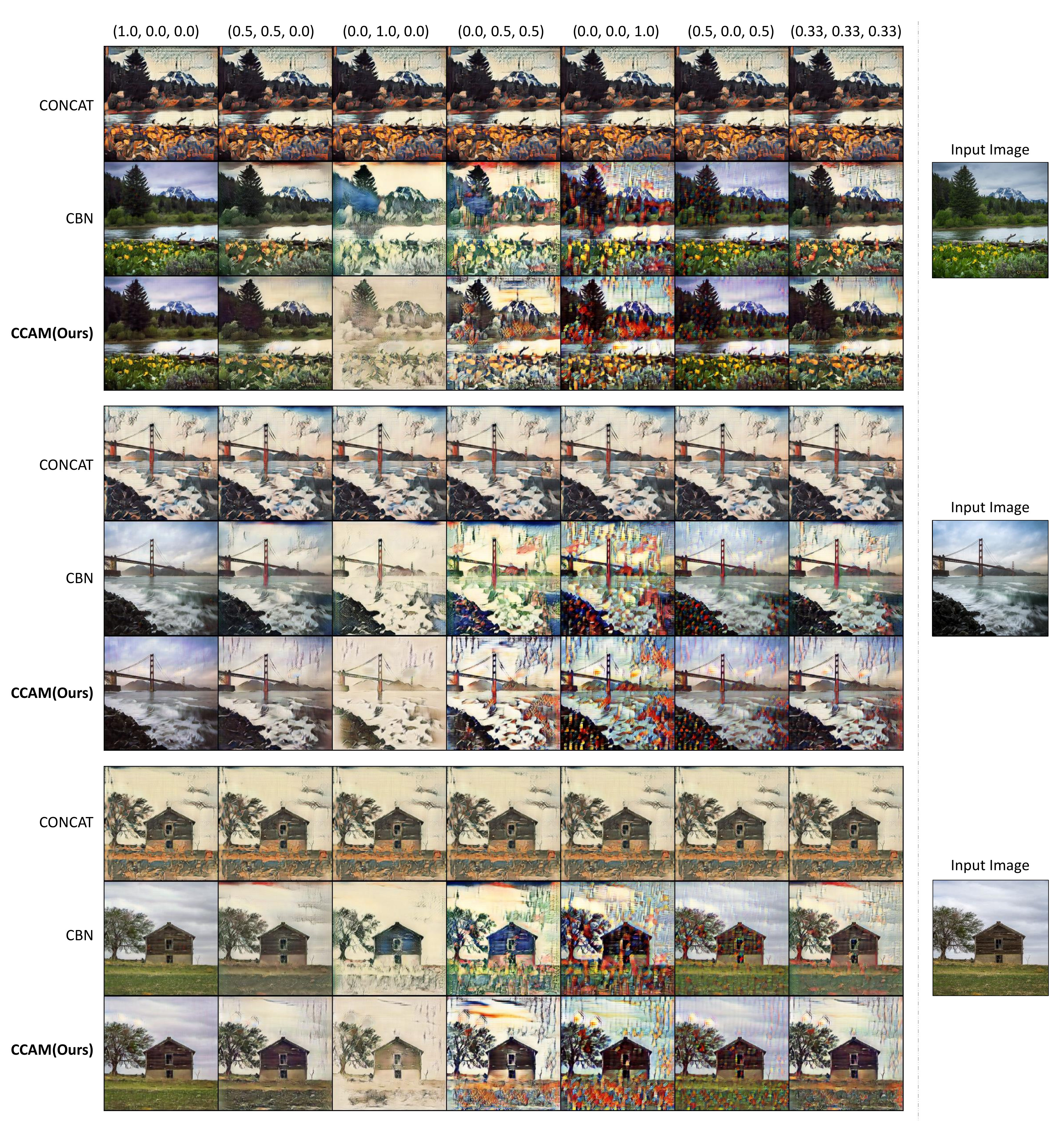}
    \caption{\textbf{More results on injection methods} All settings except for the sym-parameter injection method are equivalently set up as Model 2. CONCAT does not show any difference according to the sym-parameter, CBN shows comparatively domain characteristic, but inter-domain interference is more than CCAM. For example, if we look at the image of $S = (0.5, 0.5, 0.0)$ in CBN, intense color appears at the top, which is not related to domains A and B at all. This result shows that CBN is hard to exclude the effect of domains not related to the input sym-parameter. The CCAM used for SGN has the most explicit domain-to-domain distinction and the least impact of irrelevant domains.}
    \label{fig:injection_more}
\end{figure}

Figure \ref{fig:ch_act} summarizes channel activation trends of CCAM when changing sym-parameters for a given test image. 
As can be seen in the plots, channels are activated differently for three cases of sym-parameter.
First layer of CCAM is mainly responsible for scaling with no blocked channel. 
Considering the number of closed channels with zero activations are increased at deeper layers, CCAM selectively excludes channels and reduces influence of unnecessary channels to generate images in a mixed-domain conditioned by sym-parameters.
This is the major difference of our CCAM and the CBN which utilizes bias in controlling each domain's influence.
Figure \ref{fig:injection_more} is additional images according to the sym-parameter injection method.

\subsection{More Results of Image Translation}
\begin{figure}
    \centering
	\includegraphics[width=1.\linewidth]{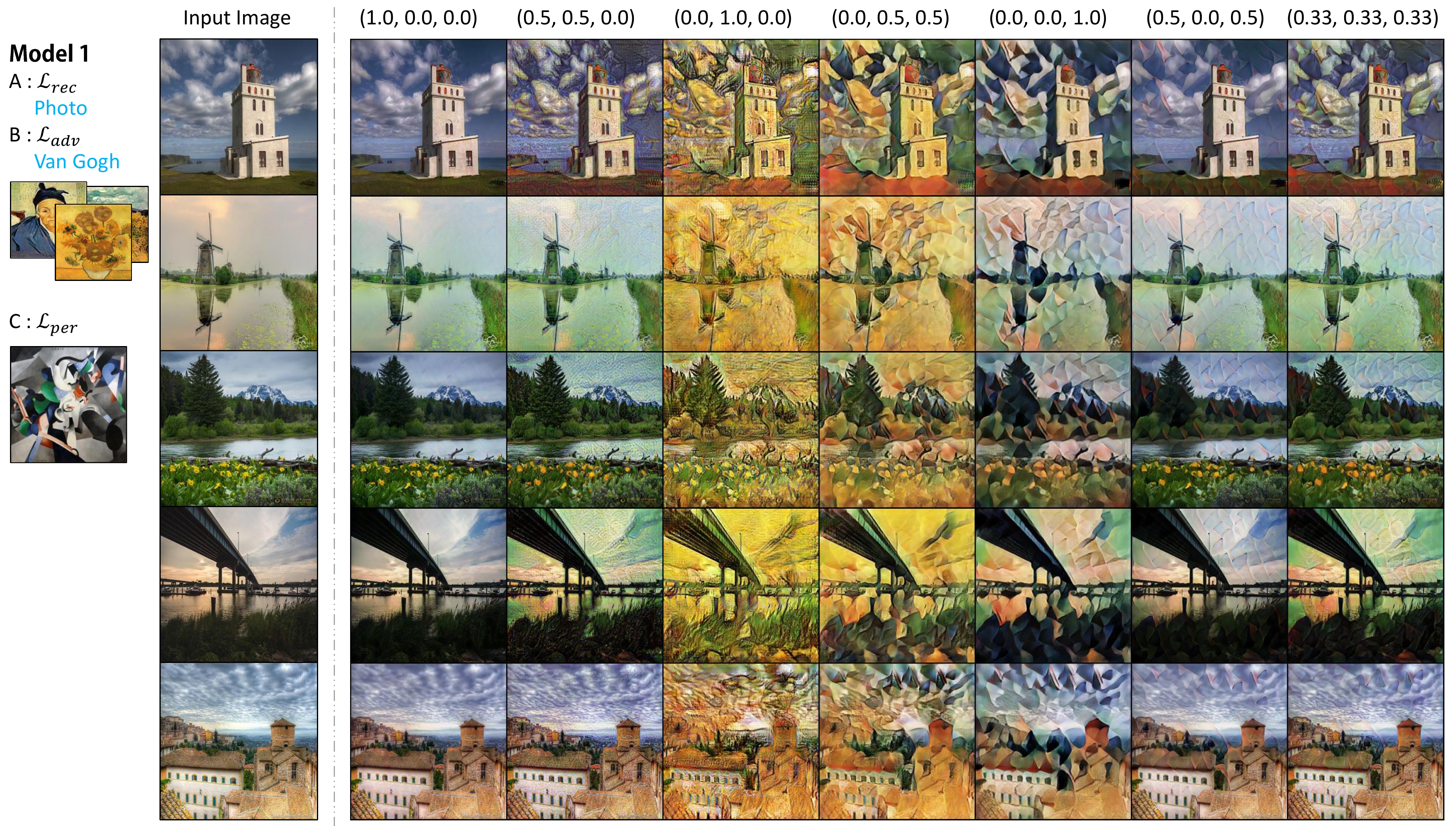}
    \caption{\textbf{More results of SGN Model 1.} The
numbers in the parentheses are sym-parameters for each A, B, and C domain.}
    \label{fig:model1}
\end{figure}

\begin{figure}
    \centering
	\includegraphics[width=1.\linewidth]{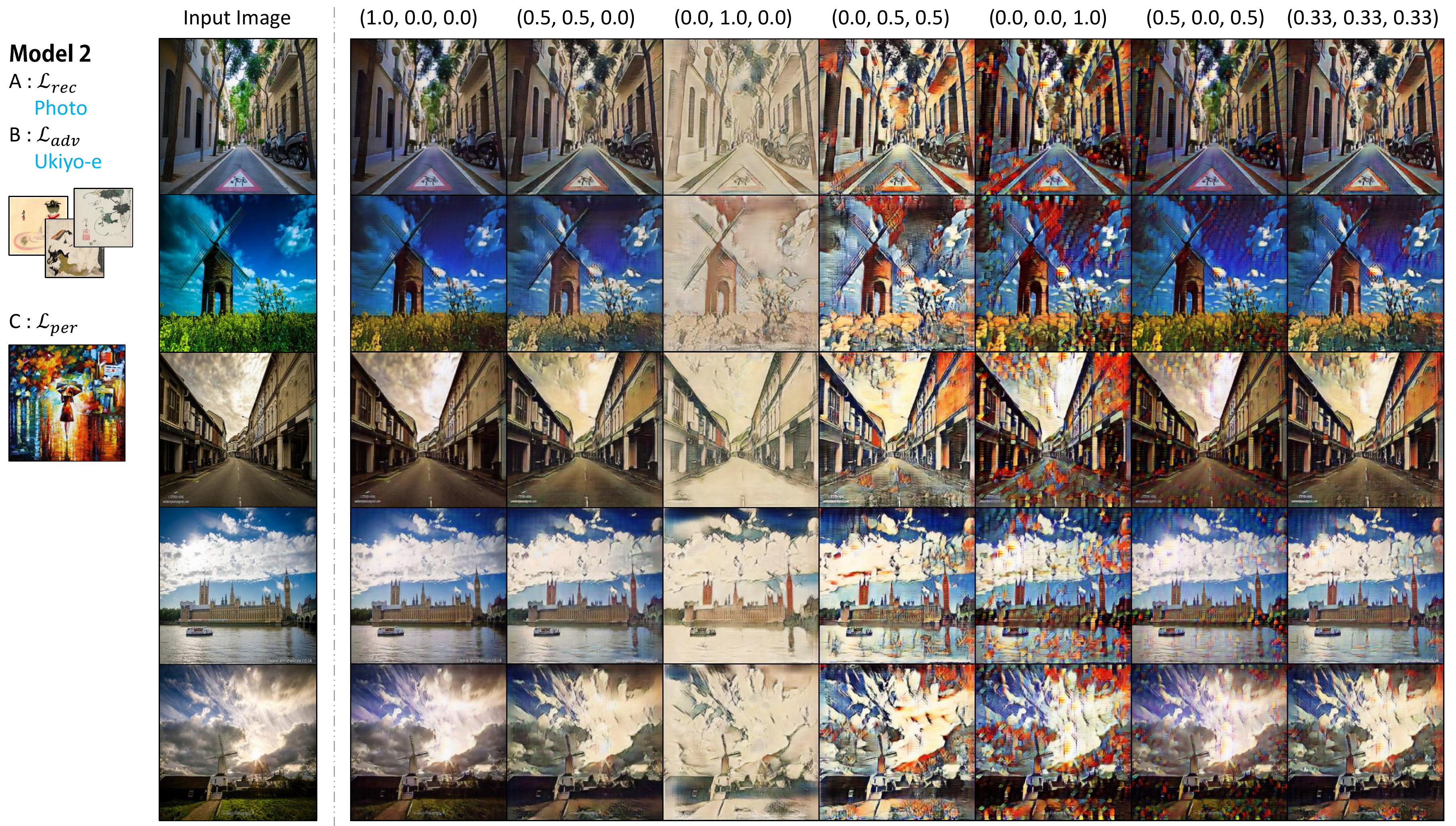}
    \caption{\textbf{More results of SGN Model 2.} The
numbers in the parentheses are sym-parameters for each A, B, and C domain.}
    \label{fig:model2}
\end{figure}

\begin{figure}
    \centering
	\includegraphics[width=1.\linewidth]{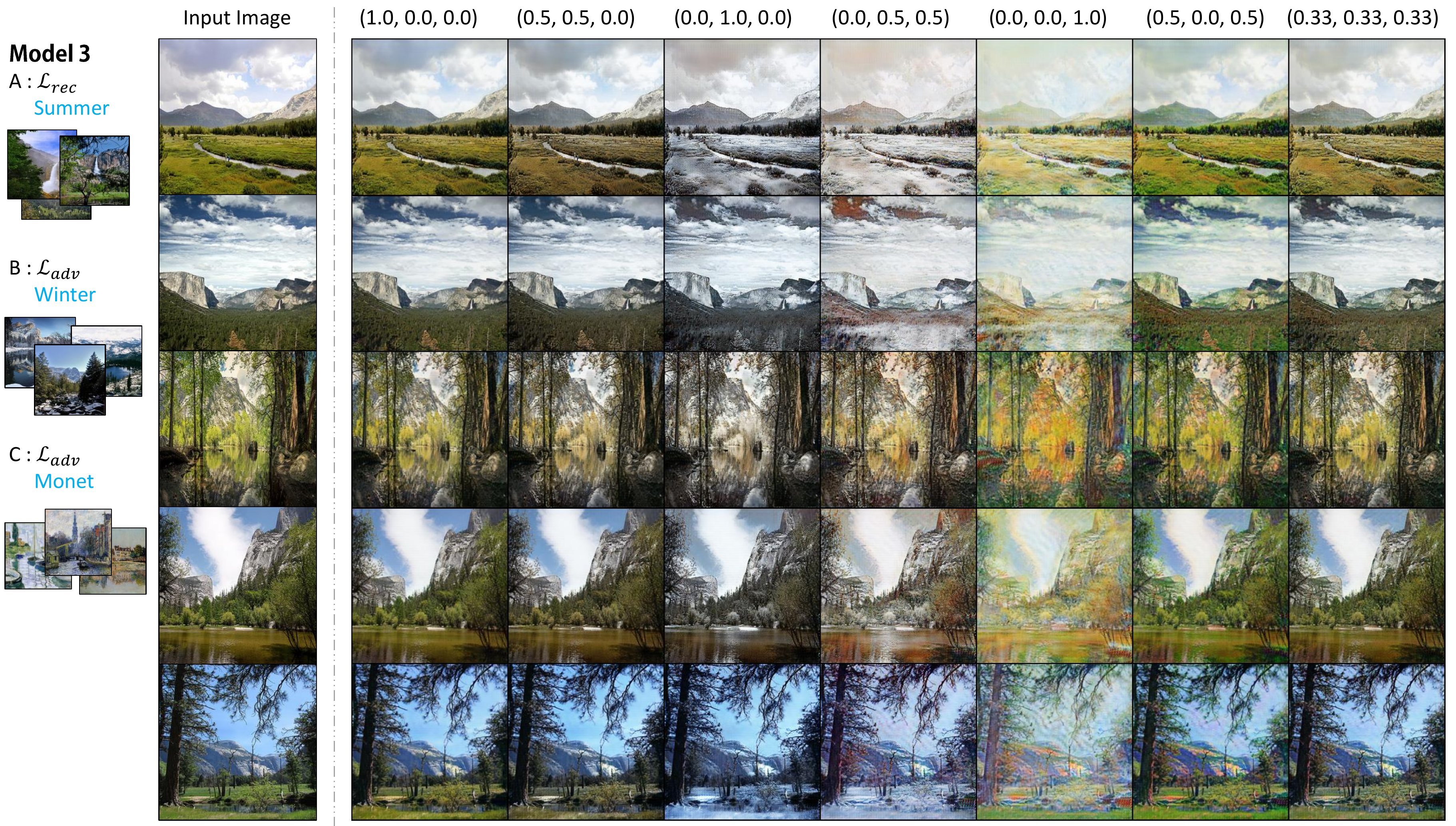}
    \caption{\textbf{More results of SGN Model 3.} The
numbers in the parentheses are sym-parameters for each A, B, and C domain.}
    \label{fig:model3}
\end{figure}

\begin{figure}
    \centering
	\includegraphics[width=1.\linewidth]{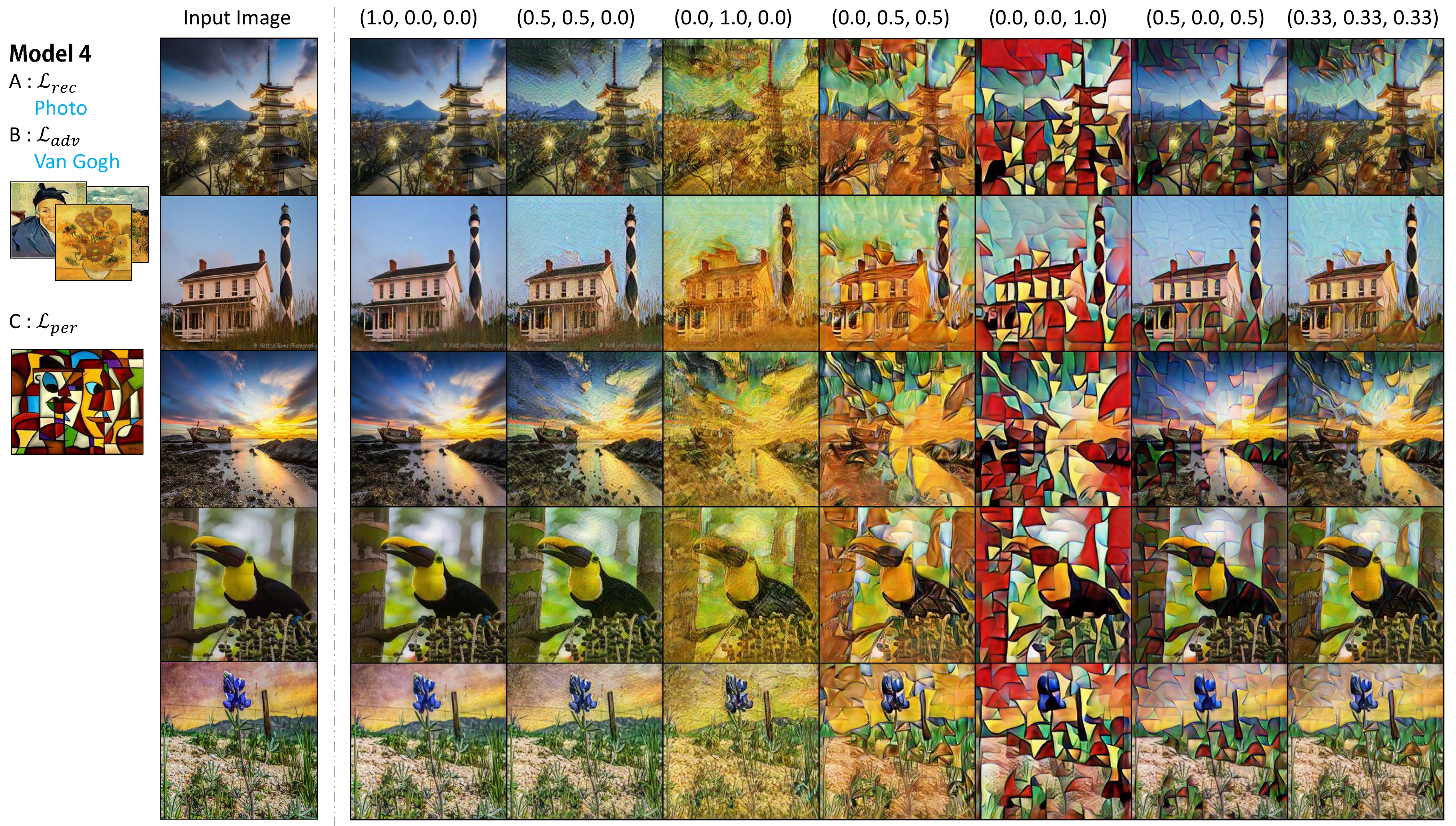}
    \caption{\textbf{Translation results of additional SGN model.} The
numbers in the parentheses are sym-parameters for each A, B, and C domain.}
    \label{fig:model4}
\end{figure}

Figure \ref{fig:model1}, \ref{fig:model2} and \ref{fig:model3} shows additional images for models 1, 2, and 3 of the paper, respectively. Figure \ref{fig:model4} is the results of the new SGN model not in the paper.

\end{document}